\documentclass[10pt,journal,compsoc]{IEEEtran}

\ifCLASSOPTIONcompsoc
  \usepackage[nocompress]{cite}
\else
  \usepackage{cite}
\fi
\ifCLASSINFOpdf
\else
\fi
\usepackage{tabularx}
\usepackage{booktabs}
\usepackage[hidelinks]{hyperref}
\usepackage[ruled,vlined]{algorithm2e}

\usepackage{subfigure}
\usepackage{amsmath}
\usepackage{graphicx}
\usepackage{caption}
\begin{document}

%
\title{A Survey Of Regression Algorithms And Connections With Deep Learning}
%
%
%
%

\author{Yunpeng Tai~\IEEEmembership{}
  \IEEEcompsocitemizethanks{\IEEEcompsocthanksitem Yunpeng Tai is a freshman of  Suzhou University of Science and Technology.\protect\\
    E-mail: yunpengtai@foxmail.com.
  }}

\IEEEtitleabstractindextext{%
  \begin{abstract}
    Regression has attracted immense interest lately due to its effectiveness in tasks like predicting values. And Regression is of widespread use in multiple fields such as Economics, Finance, Business, Biology and so on. While considerable studies have proposed some impressive models, few of them have provided a whole picture regarding how and to what extent Regression has developed. With the aim of aiding beginners in understanding the relationships among different Regression algorithms, this paper characterizes a broad and thoughtful selection of recent regression algorithms, providing an organized and comprehensive overview of existing work and models utilized frequently. In this paper, the relationship between Regression and Deep Learning is also discussed and a conclusion can be drawn that Deep Learning can be more powerful as an combination with Regression models in the future.
  \end{abstract}
  
  }

\maketitle

\IEEEdisplaynontitleabstractindextext

%
\IEEEpeerreviewmaketitle

\ifCLASSOPTIONcompsoc
  \IEEEraisesectionheading{\section{Introduction}\label{sec:introduction}}
\else
  \section{Introduction}
  \label{sec:introduction}
\fi

%
%
%
%
\IEEEPARstart{R}{egression} is an approach of obtaining a relationship between input space and output space. The relationship is represented by a function  $f:X\longmapsto Y$ and $X$ is known as the independent variable, and $Y$ as the dependent variable. It  was originally put forward by Legendre \cite{Legendre} in 1805, who applied least squares in Regression. And then Gauss \cite{Gauss} published a further development of the theory of least squares featuring ordinary least squares in 1821. 

Regression belongs to supervised learning and $Y$ is continuous, i.e. $Y \in R$. Undoubtedly, Regression is powerful and has made tremendous impact in enormous fields. As such, an increasing number of research has made fundamental improvement to Regression models over the past few decades. 

There has been such a surge of Regression models proposed recently, that researchers and beginners may find it challenging to figure out what exactly every model means and the relationship between them. Thus, a survey of the existing Regression models is beneficial both to beginners who just want to scratch the surface of Regression and researchers willing to have a systematic view of Regression models and gain insight from those smart models.

The key component of Regression is ordinary least squares. It is capable of producing an unbiased linear model of minimum variance as long as six necessary assumptions is satisfied according to Gauss-Markov Theorem \cite{Gauss}. However, if OLS is applied in specific areas, some of the assumptions are likely to be violated so that OLS fails to play its part in predicting values. Hence, it is essential to grasp those assumptions and figure out possible solutions when one of them is broken. And those well-known concerns with OLS contributes to extensive models designed to fix those violations such as Ridge \cite{Ridge_1}\cite{Ridge_2}, Lasso \cite{Lasso}, Elastic Net \cite{Elastic} and so on.

What distinguishes this paper from others is the earlier part of this paper is OLS-centered and alternative solutions provided by different models are discussed at length when some assumptions are broken (Figure 1). This paper views the relationship between models on the whole and discusses the details of distinct models specifically and explicitly. And this paper also provides a walk-through of some uncommon Regression models. What's more, a possible direction to which Regression is going to develop is covered.

 \begin{figure}[h]
\includegraphics[height=2in]{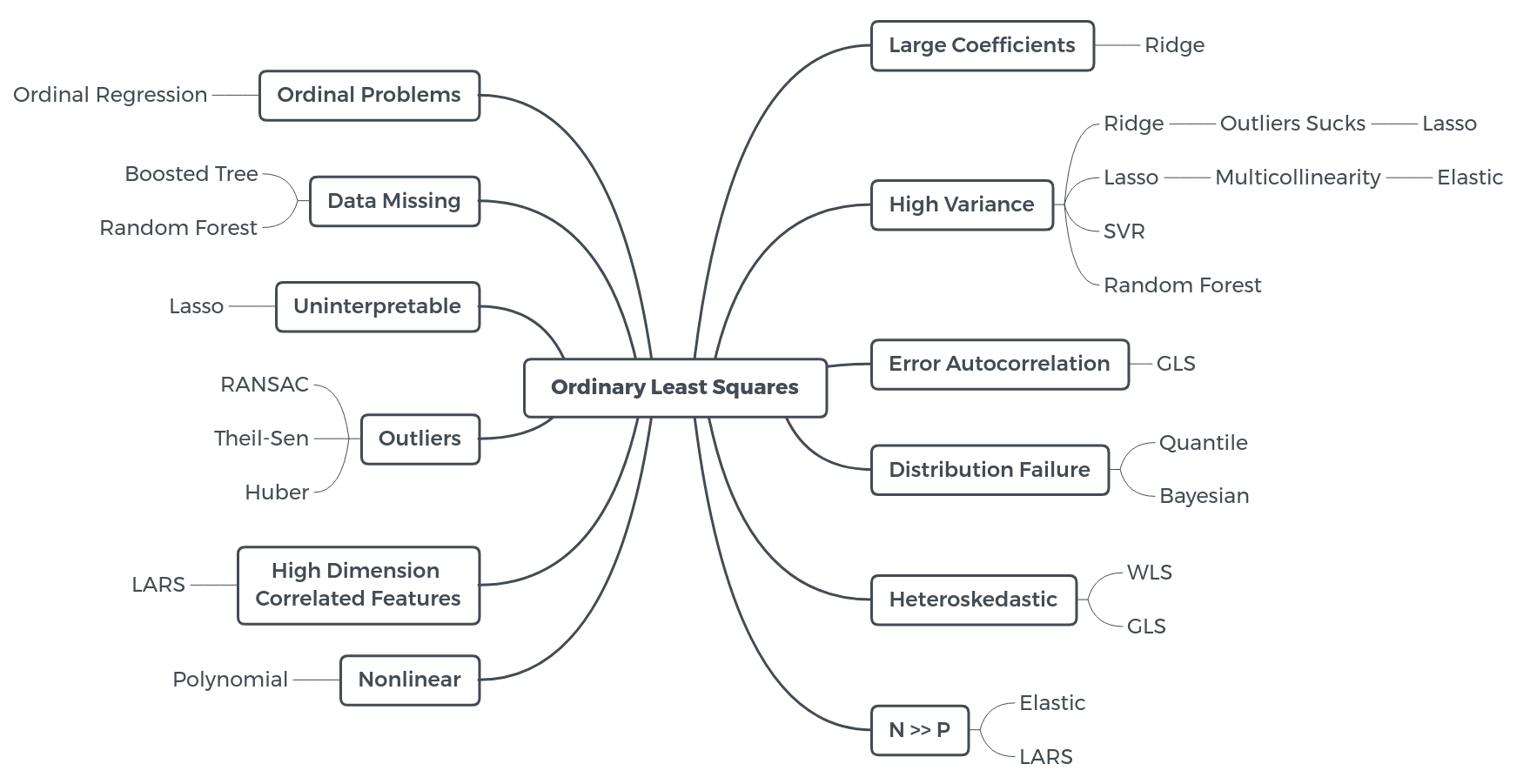}
\caption{Problems With OLS And Possible Solutions 
Provided By Distinct Models. 
N is the size of input space. 
P is the number of features of every sample in input space.}
\end{figure}

	\begin{figure*}[t]
		\center
		
		\subfigure[$X_T=log(x)$]{\label{fig:1a}
		\includegraphics[width=0.15\linewidth]{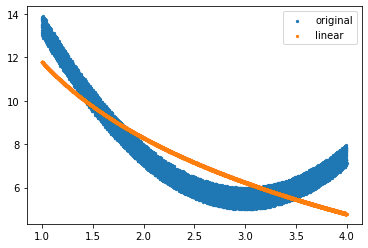}}
		\subfigure[$X_T = \sqrt{x}$]{\label{fig:1b}
		\includegraphics[width=0.15\linewidth]{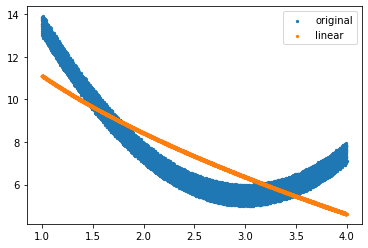}}
		\subfigure[$X_T = exp(x)$]{\label{fig:1c}
		\includegraphics[width=0.15\linewidth]{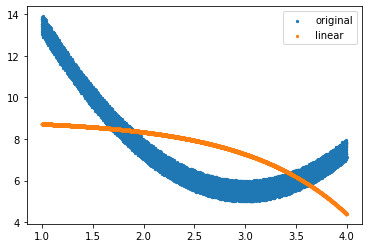}}
		\subfigure[$X_T = \frac{1}{x}$]{\label{fig:1d}
		\includegraphics[width=0.15\linewidth]{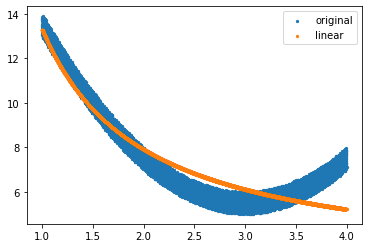}}
		\subfigure[$X_T = (x-3)^2$]{\label{fig:1e}
		\includegraphics[width=0.15\linewidth]{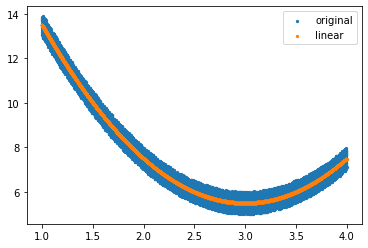}}

		\center
		\subfigure[$X_T=x^3$]{\label{fig:2f}
		\includegraphics[width=0.15\linewidth]{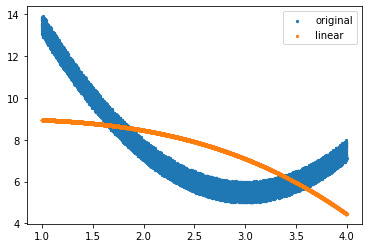}}
		\subfigure[$X_T =x^4$]{\label{fig:2g}
		\includegraphics[width=0.15\linewidth]{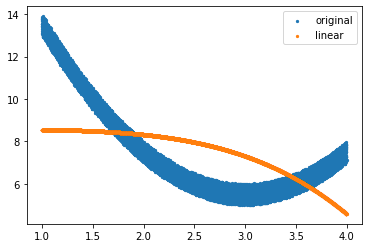}}
		\subfigure[$X_T = x^5$]{\label{fig:2h}
		\includegraphics[width=0.15\linewidth]{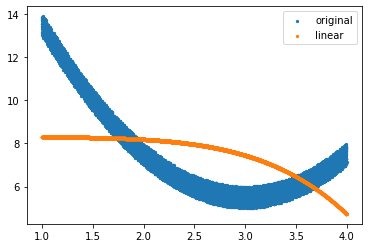}}
		\subfigure[$X_T = x+1$]{\label{fig:2i}
		\includegraphics[width=0.15\linewidth]{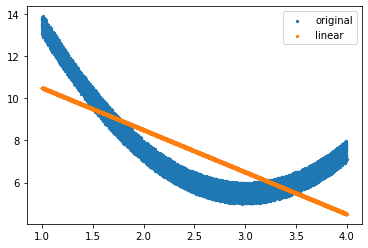}}
		\subfigure[$X_T = \frac{1}{x^2}$]{\label{fig:2j}
		\includegraphics[width=0.15\linewidth]{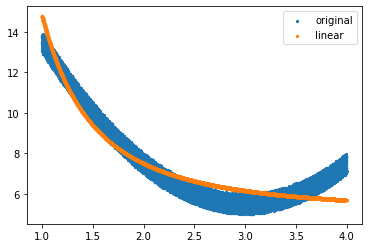}}
	
     	\centering	
		\caption{Visualize Transformed Results. Every figure corresponds to a function.}

	\end{figure*}

This paper is organized as follows. A brief introduction of the Regression task and convention in this paper is included in Section 2. In Section 3, OLS, its assumptions and possible solutions for violation are comprehensively explained. Section 4 is composed of a number of Regression models which enable OLS's potential to be stimulated although the real data challenges its assumptions and some unexpected situations happen. Generalized Linear Models and an uncommon Regression named as Step-Wise are explored in Section 5. Section 6 sets out to provide a quick overview of the strong bond between Regression and Deep Learning. In Section 7, conclusions about Regression are drawn and possible combination of Regression and Deep Learning in the future is discussed. In a nutshell, this well-established paper is an overview of Regression models and the relationship between Regression and Deep Learning, and hopefully this paper does make sense.

\section{The Regression Task}
Our data looks like $\left\{(x_1,y_1),(x_2,y_2),...,(x_n,y_n)\right\}$, which $y_i  \in R$. We intend to train a model from our data set and implement it in unknown test sets. A standard for machine performs well is that low residuals(distance from predicted values to labels). When it comes to regression task, it is common way to implement Linear Regression.
\begin{subequations}
\renewcommand{\theequation}{\theparentequation.\arabic{equation}}
\begin{align}
y &= \hat{y} + \epsilon \\
  &= \theta x+ b + \epsilon 
\end{align}
\end{subequations}


$\theta$ is called slope(gradient) or coefficient and $b$ is called intercept. $\theta$ explains when $x$ changes to what extent $\hat{y}$ is going to change. $X = \left\{x_1,x_2,...,x_n\right\}$ is known as input space and  $Y = \left\{y_1,y_2,...,y_n\right\}$ as output space. $x_i$ is called a sample, and $x_{ij}$ means the j-th feature of the i-th sample. $y$ is the label and $\hat{y}$ is the prediction. And $\epsilon$ is the error accompanied by every prediction and is also called the distance from $\hat{y}$ to $y$(residual). In ordinary least squares, the model assumes that $y$ is actually sampled from Gaussian Distribution and every sample is with noise. Thus, it can also called noise in statistics. But in this paper, I am going to use error for that.

\section{Ordinary Least Squares}
In Machine Learning, we always figure out the best model by minimizing our objective function, which is also known as cost function. OLS(Ordinary Least Squares) serves as an effective loss function as long as the model satisfies six necessary assumptions. Then it can choose an unbiased model of minimum variance by minimizing the function below. And $J(\theta)$ is convex. Thus, set partial derivative of $\theta$ zero and then we can get the best parameter $\theta^*$. Note that only when $X^TX$ is full rank, equation(3) does make sense. Some books may multiply $J(\theta)$ by $1 / n$, which is convenient for computation. Note that $X$ and $Y$ are matrices. 
\begin{align}
    J(\theta) &=  \sum \limits_{i} ^ {n} (y_i - (\theta x_i + b))^2\\
    \theta ^ * &= (X^TX)^{-1}X^TY
\end{align}

\subsection{Prior Assumptions}

In this section, six necessary assumptions is studied. And possible solutions for unsatisfied situations are also covered.

\begin{itemize}
\item  Linearity. In other words, only Straight Line models are permitted. If the relationship between $X$ and $y$ is a non-linear model, e.g. $y = X^4$, the whole regression model crashes.
And the recipe for this situation is applying feature transformation. By doing so, the whole relationship between $X$ and $y$ is changed for good. Hence, we must also take the correlation between $X$ and $y$ into consideration. Correlation can be told by calculating $R^2$ (Coefficient Of Determination), which stands for the ability to predict $y$ by observing $X$. When $R^2 = 1$, it means the loss of this predictor is 0. If $R^2 = 0$, it is equivalent to that the predictor is constant, which indicates $X$ has nothing to do with $y$. As shown in Table 1, in Column $R^2$, the value in bracket stands for the original coefficient. And the same goes for Column Linearity. Note that I use 10000 random points from $y = 2 (X - 3 ) ^ 2 + 5 + N$, which N stands for noise and it varies from 0 to 1. X follows random and even distribution. 
\begin{align}
        \overline{y} &= \frac{1}{n}\sum\limits_{i=1}^{n}y_i\\
                        R^2 &= 1 - \frac{\sum\limits_{i=1}^{n}(y_i - \hat{y_i})^2}{\sum\limits_{i=1}^{n}(y_i-\overline{y})^2}
\end{align}
	\begin{figure}[t]
		\center
		\subfigure[Errors]{\label{fig:1a}
		\includegraphics[width=0.3\linewidth]{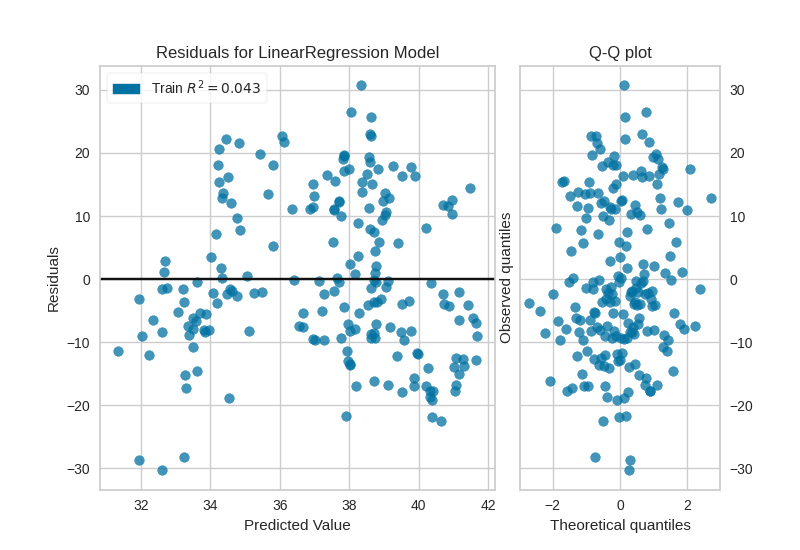}}
		\subfigure[Norml Distribution]{\label{fig:1b}
		\includegraphics[width=0.3\linewidth]{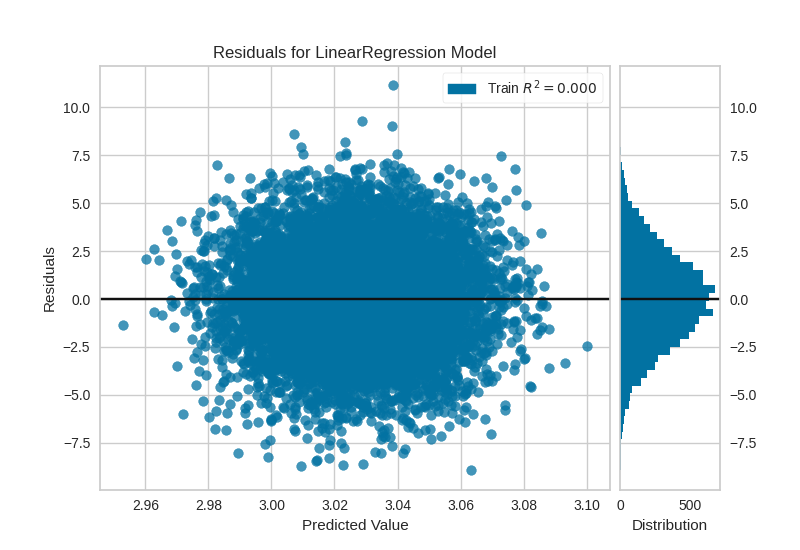}}
		\subfigure[Q-Q of ND]{\label{fig:1c}
		\includegraphics[width=0.3\linewidth]{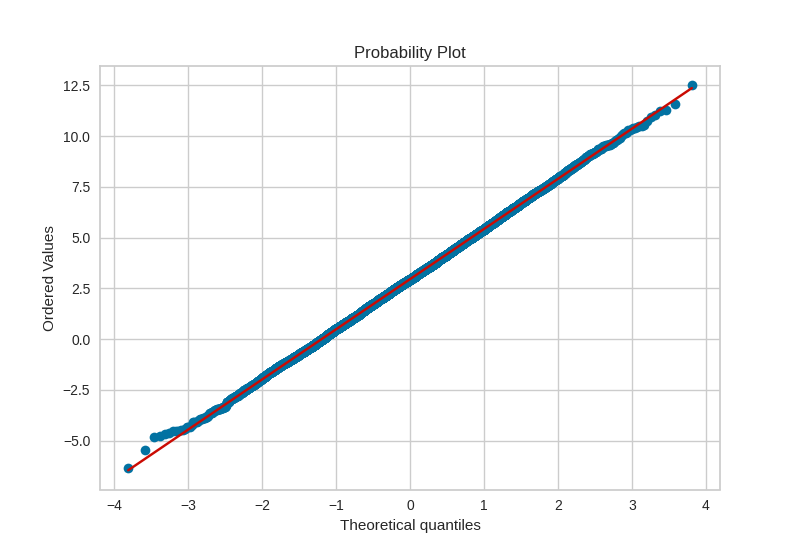}}
	\caption{In Figure(a), errors are not symmetrical and Q-Q plot doesn't look like a line,which indicates errors don't follow Normal Distribution. Figure(b) and Figure(c) show when errors follow Normal Distribution, the hist plot should look like bell curve and Q-Q plot should be a line.}

	\end{figure}

\begin{table}[!h]
\centering
           \caption{Transformation Results About $R^2$ And Linearity}
\begin{tabular}{lllll}
\toprule  
 $X$ & $y$	& $Transformation$&$R^2$(-1.602) & $Linearity(Non)$\\
\midrule  
1.450 & 9.994    & $X_T = log_{10}(x)$&	0.764 & Non\\
1.003 &	13.475  & $X_T = \sqrt{x}$	& 0.693 & Non\\
3.179 & 5.935	 & $X_T =  exp(x)$&	0.281 & Non\\
3.801 &	6.726    & $X_T = \frac{1}{x}$ & 0.870 & Non \\
1.294 & 11.704  & $X_T = (x - 3)^2$& 0.983 &	Linear\\
1.399 &	10.915  & $X_T = x^3$ &0.344 & Non\\ 
3.591 & 5.951    & $X_T = x^4 $&0.253 & Non\\
2.987 & 6.742	 & $X_T = x^5$&0.188 & Non\\
3.161 & 5.906    & $X_T = x + 1$& 0.616 & Non\\
2.065 & 7.647    & $X_T  = \frac{1}{x^2}$& 0.904 & Non\\
\bottomrule 
  
 \end{tabular}

 \end{table}

As shown in Table 1, $R^2$ changes when feature transformation is applied. And the purpose is to find linear model with the best $R^2.$ Note that linear means y w.r.t transformed X. And if  so, scatter plot of original X and predicted y should fit the original distribution plot. This assumption is the most significant for Linear Regression and it may explain why Ridge or Lasso also performs badly when the relationship is nonlinear.
\end{itemize}

\begin{itemize}
\item Constant Error Variance \cite{Hetero}. It means errors are uniformly distributed, which in statistics is called no Heteroscedasticity. When we apply our model, we can get a bunch of predicted values via observing $X$. Then we can calculate errors between true values and predicted values. And we can also calculate the variance of errors. If errors follow normal distribution (equation 5), thus its variance is constant($\sigma^2$). Also, its distribution is symmetrical. In turn, the distribution of errors should also be uniform and symmetrical. So we can use error plot to detect it (Figure 3). And Q-Q Plot can detect whether the errors follow normal distribution. The data I use can be downloaded \href{https://www.kaggle.com/quantbruce/real-estate-price-prediction}{here}. I choose X2 house age as $X$ and Y house price of unit area for $y$. Note that I remove points that $X$ equals $0$. And I choose 200 points for study.  

\begin{align}
    f(x) &= \frac{1}{\sigma\sqrt{2\pi}}e^{-\frac{1}{2}(\frac{x-\mu}{\sigma})^2}\\
    f(x) &= \frac{1}{\sqrt{2\pi}}e^{(-\frac{x^2}{2})}
\end{align}
equation(6) represents standard normal distribution, $\mu = 0, \sigma = 1$. As shown in Figure 3, if errors follow normal distribution, the distribution should be uniform just like Figure (b) and histogram should be like bell curve. We can also draw a safe conclusion that if errors follow normal distribution, their distribution should fit the line in Q-Q Plot. 
	
           \begin{figure}[t]
		   \center
		   \subfigure[Before Log]{\label{fig:1a}
		   \includegraphics[width=0.4\linewidth]{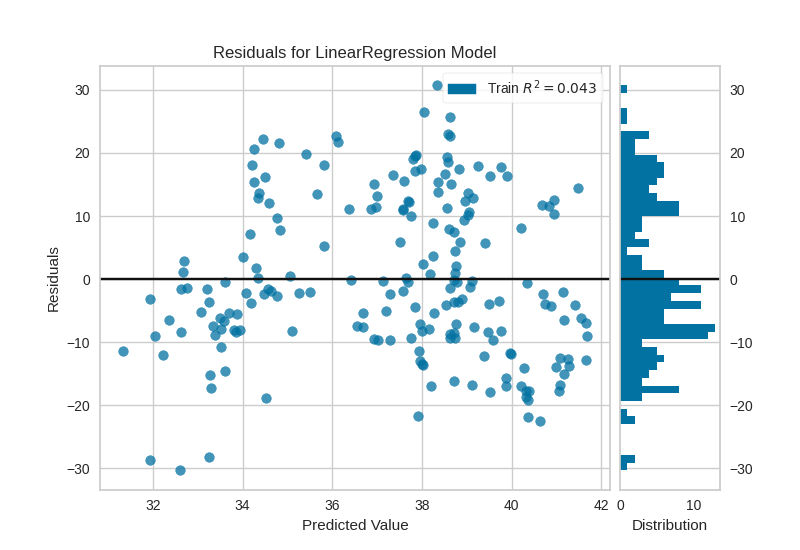}}
		   \subfigure[After Log]{\label{fig:1b}
		   \includegraphics[width=0.4\linewidth]{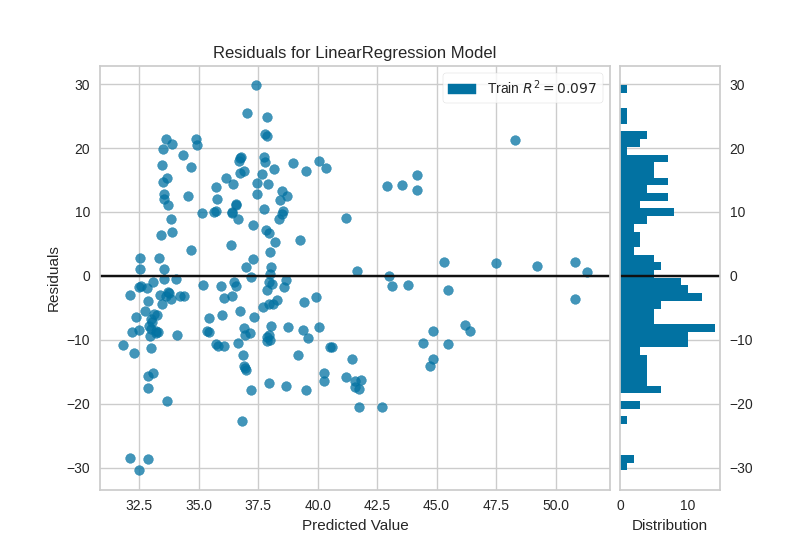}}
      \caption{The Relationship Between Residuals And Log}

        	\end{figure}
After figuring out heteroscedasticity, we can come up with a question that how it influences our model and how to improve it. The most common way is to try feature transformation e.g. Log. As shown in Figure 4, it can, to some degree, make our errors' distribution slightly more stable. It's always an option to try, but not an effective method to handle the problem.

What's more, we can apply Box Cox Transformation, which can make data more close to normal distribution. In statistics, if data follows normal distribution and then the variance of noise(error) is a constant($\sigma^2$). Thus,normality of data is likely to relieve heteroscedasticity. And data can be downloaded \href{https://archive.ics.uci.edu/ml/machine-
 learning-databases/wine-quality/winequality-white.csv}{here}. I choose total sulfur dioxide for $X$and quality for $y$. In Figure 5, it may relieve heteroscedasticity. Note you can't always depend on it. Also, it can be worse(Figure 6).

\end{itemize}

\begin{itemize}
\item Independent Errors(no autocorrelation, AC for short). For instance, you want to predict the shares in stock market. But the errors are correlated while they should be $i.i.d$(independent identically distributed). When a financial crisis happens, the shares is going to be extremely unstable in next few months,which means errors are going to increase sharply. It can be detected by Durbin Watson Test(Table 3) or drawing AC Plot. And if values of y axis in AC Plot vary from $(0,1]$, they mean Positive AC. If values equal 0, they mean Non AC. Otherwise, they mean Negative AC.

\begin{table}[!h]
\centering
\setlength{\belowcaptionskip}{0.1cm}
\caption{Durbin Watson Test}
\begin{tabular}{ll}
\toprule  
 $Value$ & $Relationship$\\
\midrule  
2.0    & no Ac\\
0.0-2.0& positive AC	\\
2.0 - 4.0 & negative AC\\
\bottomrule 
  
 \end{tabular}
 \end{table}
 The autocorrelation can affect errors' standard deviation while it's unlikely to have an influence on model's coefficient and intercept \cite{autocorrelation}. 

There're two common ways to fix it. The first is to add omitted variables. For example, you want to predict stock performaces by time. Undoubtedly, the model is of hight autocorrelation. We can, however, add S\&P 500. And hopefully, it may relieve autocorrelation. The second is to switch another function. You can transform your linear model into a squared model. 
\end{itemize}

\begin{itemize}  

	\begin{figure}[htbp]
	\item
		\center
		\subfigure[Density Distribution]{\label{fig:1a}
		\includegraphics[width=0.3\linewidth]{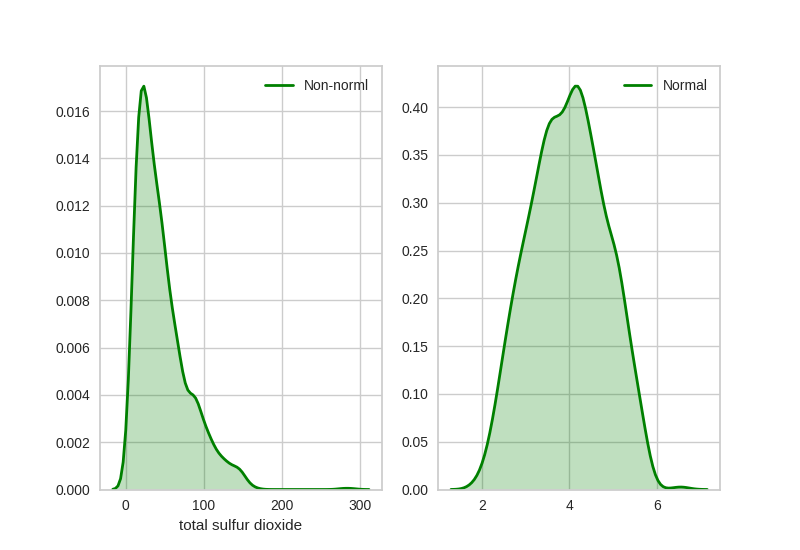}}
		\subfigure[Original Residual]{\label{fig:1b}
		\includegraphics[width=0.3\linewidth]{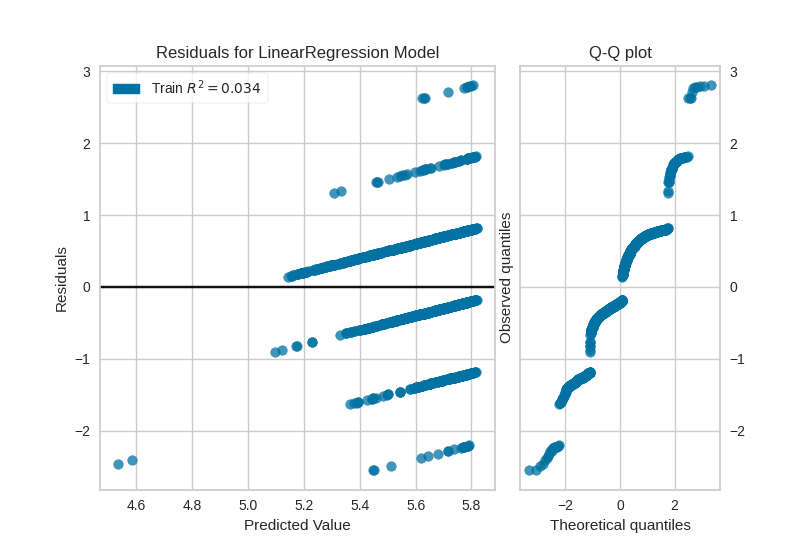}}
		\subfigure[Final Residual]{\label{fig:1c}
		\includegraphics[width=0.3\linewidth]{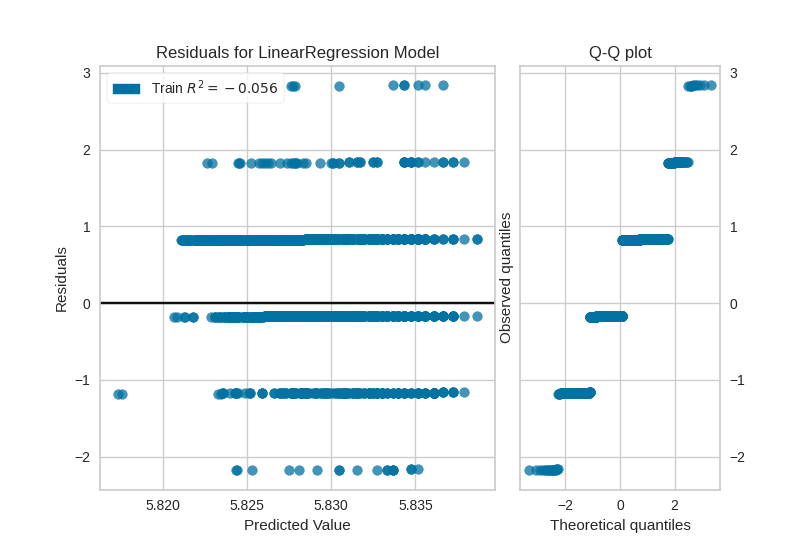}}
	\caption{Box Cox Transformation May Help}

	\end{figure}

\item No Multicollinearity. If independent variables are related to each other, there's Multicollinearity in data. We can use Variance Inflation Factor(VIF) to detect it($R^2$ is Coefficient Of Determination). If value = 1, it implies that there is no Multicollinearity among the predictors. If value \textgreater 5, it implies there's potential Multicollinearity. If value \textgreater 10, it implies apparent Multicollinearity.

\begin{equation}
          softmax(x) = \frac{e^{x_i}}{\sum_{i=1}^{n}e^{x_i}}
\end{equation}
Our goal of Regression model is to figure out the relationship between independent variable(X) and dependent variable(y) by finding a proper coefficient. But when there's Multicollinearity, the coefficient is unable to interpret. We actually don't know what exactly the relationship is. However, if we just want to make good predictions, it's still effective \cite{Hetero}. And if the degree of Multicollinearity is moderate, you don't have to care about it too much. We can remove highly correlated variables or increase sample size.
\end{itemize}

\begin{itemize}
\item Normality Of Data \cite{Hetero}. Box-Cox is the efficient transformation to make data more close to normal distribution. Normalization and some basic feature transformation may help. And also try increasing data size.
\end{itemize}

\begin{itemize}
\item No Exogeneity. If $X$ we choose iteself is of little influence on $y$, which means the real prediction is not based on $X$, then there's exogeneity \cite{Hetero}. And the best solution is to make a good analysis about what on earth affects our predicted values and choose a suitable $X$.
\end{itemize}

\section{Alternative Models }
In this section, OLS's weaknesses are going to appear in Background. And each Background stands for a specific problem.
\subsection{Ridge Regression}
\begin{itemize}
\item Background: high variance. OLS enables the predictor to perform good on training sets, however, this is an invitation to poor performance on testing sets, which is also known as overfitting. Generally speaking, the more complicated a model is, the poorer performance of the model on unknown sets. According to Occam's rule, a model is more likely to do a good job on unknown sets if the model is simple. And the model is of high generalization ability. 
\end{itemize}
\begin{itemize}
\item Shrinkage. When OLS is applied in real life, the predictor's coefficients can be too large in absolute. The coefficient of variable more correlated to $y$ is large while the one of variable less correlated to $y$ is also large,which is misleading to figure out the relationship between $X$ and $y$ \cite{Ridge_1}. This phenomenon can account for poor performance on unknown sets. Thus, Ridge is going to implement shrinkage on coefficients and the extent of shrinkage counts on the degree of correlation. Typically, if a variable holds much predicting power, its coefficient is more likely to be big \cite{Ridge_2}.
\end{itemize}
\begin{itemize}
\item Nonorthogonal Solution \cite{Ridge_1}. Whether OLS can be directly computed just by derivative relies mainly on if $X^TX$ is orthogonal. If $X^TX$ is nonorthogonal, this means  $X^TX$ is not reversible and the direct computation can't work. $I$ is a unit matrix. There're only small positive quantity on the diagonal of $kI$, the diagonal looks like a ridge in comparison with zero's distribution(equation 9).\begin{figure}[htbp]
		\center
		\subfigure[Density Distribution]{\label{fig:1a}
		\includegraphics[width=0.3\linewidth]{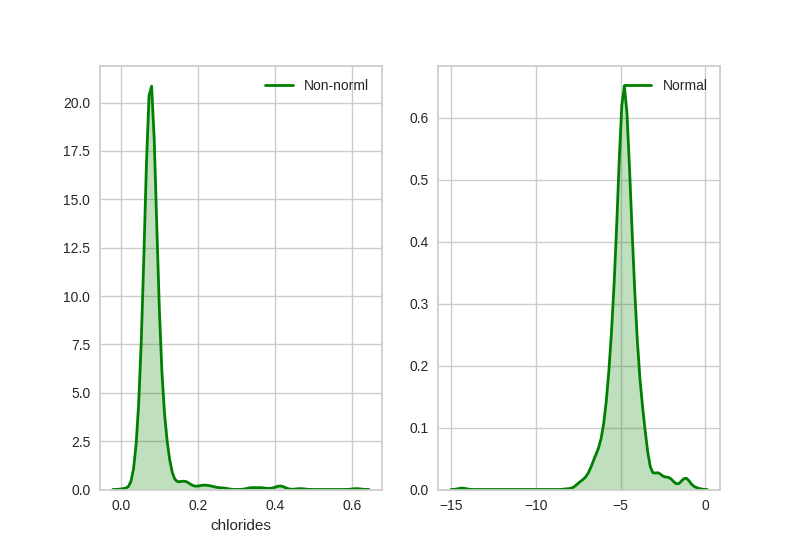}}
		\subfigure[Original Residual]{\label{fig:1b}
		\includegraphics[width=0.3\linewidth]{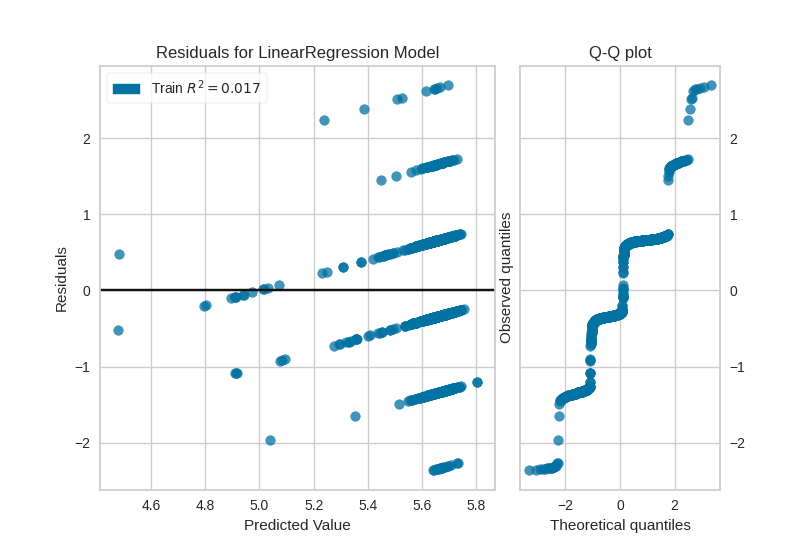}}
		\subfigure[Final Residual]{\label{fig:1c}
		\includegraphics[width=0.3\linewidth]{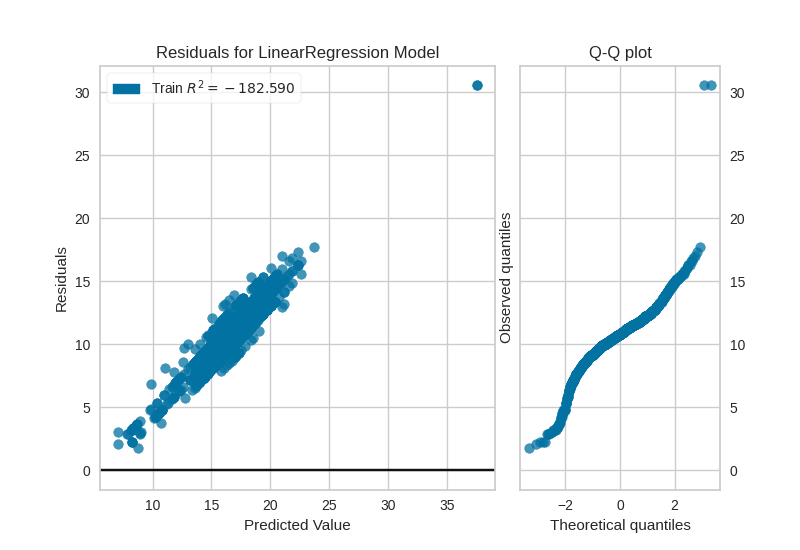}}
	\caption{Box Cox Transformation May Suck}

	\end{figure}
\end{itemize}
\begin{itemize}
\item Biased Model. The potential assumptions of Ridge are that $X^TX$ is nonorthogonal and coefficients need shrinkage. Hence, Ridge actually do a trade-off between bias and variance and it uses increased bias to obtain reduced variance. Because Ridge's assumptions are correct in most cases, it is capable of producing a model with low variance.	
\end{itemize}
\begin{itemize}
\item L2-Penalty. Equation 10 is Ridge's loss function. The loss function means not only an accurate model is required, its coefficients should be small. And $\lambda/2 ||\theta_i||^2$ is called L2-norm. This kind of method is known as regularization in the sense that the model generated by regularization is of high generalization ability. Except direct computation, gradient descent is a common way to get the best parameter. Ridge always minuses the coefficient vector(equation 11.1 \& 11.2).
\begin{equation}
\theta^* = (X^TX + kI)^{-1}X^TY
\end{equation}

\begin{equation}
        J(\theta) = \frac{1}{n}\sum \limits_{i} ^ {n} (y_i -                    \hat{y_i})^2 + \frac{\lambda}{2}||\theta _i||^2          
  \end{equation}

\begin{subequations}
\renewcommand{\theequation}{\theparentequation.\arabic{equation}}
\begin{align}
        \frac{\partial J(\theta)}{\partial \theta} &= \frac{\partial MSE}{\partial \theta} +  \lambda \theta \\
  &= \begin{bmatrix} \frac{\partial M(\theta)}{\partial \theta_1 }\\ \frac{\partial M(\theta)}{\partial \theta_2}\\ \vdots\\ \frac{\partial M(\theta)}{\partial \theta_n}\end{bmatrix} +\lambda \begin{bmatrix}\theta_1\\\theta_2\\ \vdots \\\theta_n \end{bmatrix}
\end{align}
\end{subequations}

\end{itemize}

\subsection{Lasso Regression}
\begin{itemize}
\item Background: Ridge's poor performance on outliers. Compared to OLS, Ridge is quite powerful but shrinkage means it just cuts down on coefficients' ability of affecting the result. As such, each coefficient still has influence on the result which further indicates Ridge still cares about every sample's loss. However, when outliers appear in the data, Ridge fails to deal with them in that it is sensitive to outliers.
\item Feature Selection \cite{Lasso}. Unlike Ridge, Lasso implements feature selection because only one coefficient is saved and others are set zero, which is known as sparse solution. Therefore, only one sample has effect on the prediction in the sense that Lasso is insensitive to outliers and robust to small changes. Feature selection results in oscillation in optimization, which means Ridge is more stable than Lasso in gradient descent(Figure 7). What's more, although Lasso has looked at all the data, only one sample does make sense. Thus, Lasso is also able to avoid overfitting. If performance of Lasso is excellent, then we can say which sample does work. So Lasso is an interpretable model compared to OLS and Ridge.
\begin{figure}[htbp]
\includegraphics[height=2in]{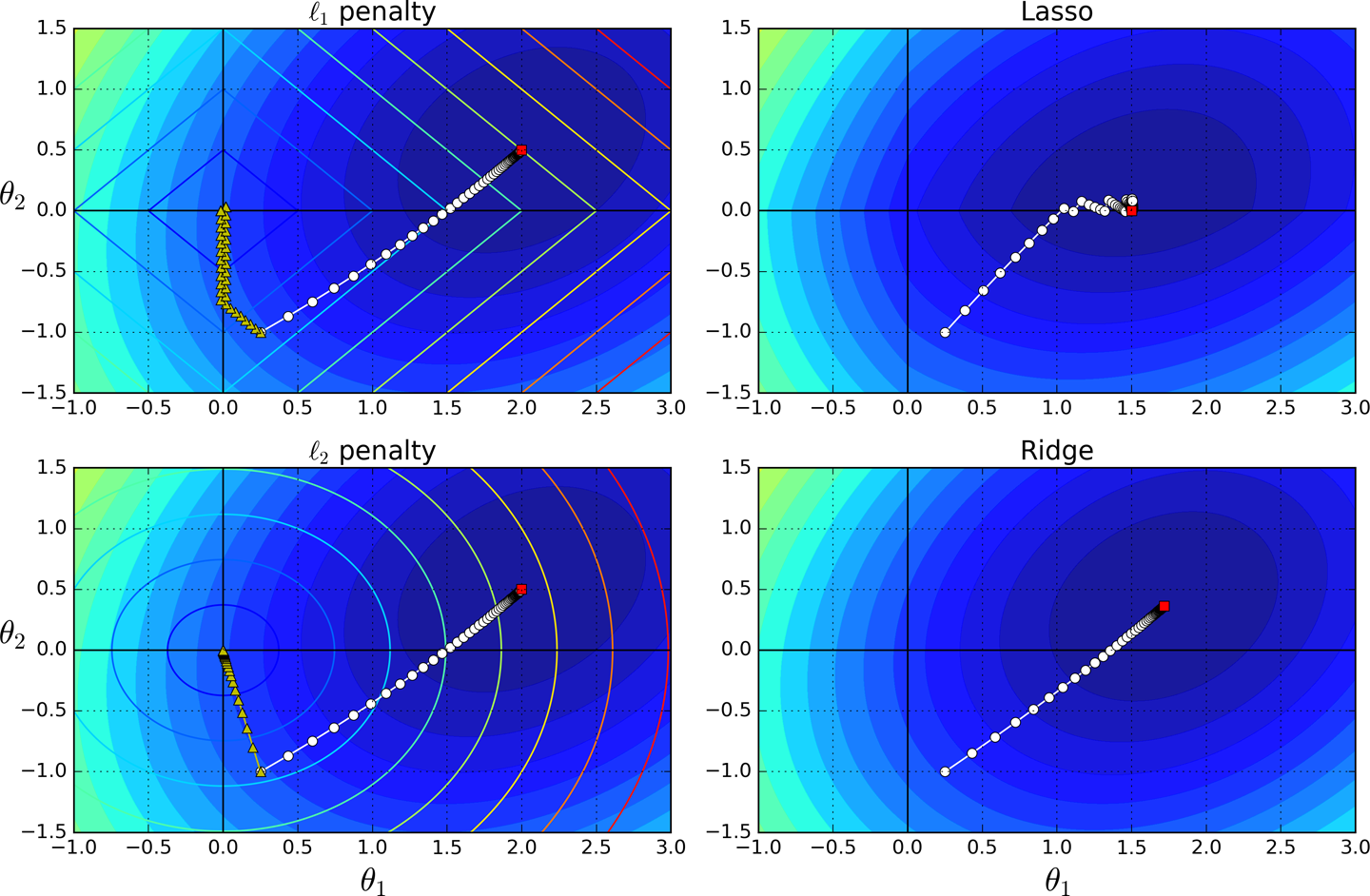}
\caption{Ridge is more stable than Lasso in optimization process(2016. Deep Learning. MIT Press).}
\end{figure}
\end{itemize}
\begin{itemize}
\item L1-Penalty.  Lasso minuses a constant in gradient descent(equation 13.1 \& 13.2). Suppose we're on the top of a mountain, what Lasso does is just move a bitter farther while Ridge just goes where seems more steap. Hence, Ridge is more faster than Lasso. So when values are quite large, Ridge should be a better choice than Lasso. But when values are small, Lasso should be a better choice.
\end{itemize}

\begin{align}
        J(\theta) &= \frac{1}{n}\sum \limits_{i} ^ {n} (y_i - \hat{y_i})^2 + \lambda||\theta_i||
\end{align}

\begin{subequations}
\renewcommand{\theequation}{\theparentequation.\arabic{equation}}
\begin{align}
        \frac{\partial J(\theta)}{\partial \theta} &= \frac{\partial MSE}{\partial \theta} + \lambda sign(\theta) \\
  &= \begin{bmatrix} \frac{\partial M(\theta)}{\partial \theta_1 }\\ \frac{\partial M(\theta)}{\partial \theta_2}\\ \vdots\\ \frac{\partial M(\theta)}{\partial \theta_n}\end{bmatrix} +\lambda \begin{bmatrix}sign(\theta_1)\\sign(\theta_2)\\ \vdots \\sign(\theta_n) \end{bmatrix}
\end{align}\end{subequations}
\subsection{Support Vector Regression}

\begin{itemize}
\item Margin Maximization. SVM(Support Vector Machine) is originally invented for classification problems \cite{svr_1}. Unlike other algorithms, SVM not only needs to classify all the data correctly but also requires the distance of data to the Hyper plane to be the biggest, which is known as widest street. Our linear model is $y = \theta ^T X + b$. Among all the data points, the distance of the closest positive point to the hyper plane pluses the same distance of closest negative point is Margin($\gamma$). Then SVM is turned into maximizing Margin. And the same for SVR. We can turn the maximizing $\frac{2}{||\theta||}$ into minimizing $\frac{||\theta||^2}{2}$. Because $||\theta||$ is bigger than 0, $||\theta||^2$ is proportional to $||\theta||$. Then the problem for SVR is like equation 15. In equation 15, C is a coefficient for regularization and $L(x)$ is an undefined loss function.
\begin{equation}
\min_{\theta,b} \, \frac{1}{2}||\theta||^2 + C \sum_{i=1}^{n} L(y_i - \hat{y_i})
\end{equation}\begin{figure}[htbp]
\includegraphics[height=1.5in]{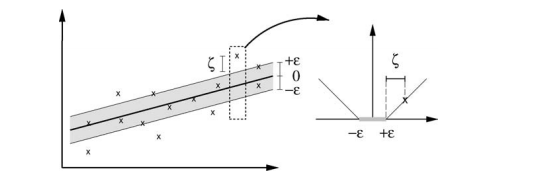}
\caption{$\epsilon$-insensitive loss. Smola and Sch$\ddot{o}$lkopf, 2002}
\end{figure}
\end{itemize}
\begin{itemize}
\item $\epsilon$- insensitive loss. SVR can tolerate mistakes which are no more than $\epsilon$, if those points predicted wrongly in dashed area $[f(x) - \epsilon,f(x) + \epsilon]$, then the losses of those points equal zero, which is called $\epsilon$-insensitive loss(Figure 8). In equation 16,z stands for loss. Actually, it just makes a trade-off between errors and complexity of the model. Hence, SVR is unlikely to overfit the data. Thus, SVR can reduce the variance of OLS.
\begin{eqnarray}
\centering
     l_{\epsilon}(z) = 
     \begin{cases}
 0 &,if \, |z|  \leq \epsilon \\
|z| - \epsilon &otherwise\\

\end{cases}
\end{eqnarray}
\end{itemize}

\begin{itemize}
\item Dual Problem. Slack variables can be introduced to optimization problem(equation 18). They means how many errors are allowed to make beyond $\epsilon$.
\begin{align}
\min_{\theta,b} \, \frac{1}{2}||\theta||^2 &+ C \sum_{i=1}^{n} L(\xi_i + \hat{\xi}_i)\\
y_i - \hat{y_i} &\leq \epsilon + \xi_i \\
\hat{y_i} - y_i &\leq \epsilon + \hat{\xi_i}\\
\xi_i \geq 0 \, &,\hat{\xi_i} \geq 0
\end{align}
We can introduce Lagrange Multiplier $\mu_i \geq 0,\hat{\mu_i} \geq 0, \alpha_i \geq 0, \hat{\alpha_i} \geq 0$. 
\begin{equation}
\begin{aligned}
&L(\theta,b,\alpha,\hat{\alpha},\xi,\hat{\xi},\mu,\hat{\mu})\\
&= \frac{1}{2}||\theta||^2 + C\sum_{i=1}^{n}(\xi_i + \hat{\xi_i}) - \sum_{i=1}^{n} \mu_i\xi_i - \sum_{i=1}^{n}\hat{\mu_i}\hat{\xi_i} \\
&+ \sum_{i=1}^{n}\alpha_i(y_i - \hat{y_i} - \epsilon - \xi_i) + \sum_{i=1}^{n} \hat{\alpha_i}(\hat{y_i} - y_i - \epsilon - \hat{\xi_i})
\end{aligned}
\end{equation}
And set partial derivative of w,b,$\xi_i$ and $\hat{\xi_i}$ zero.
\begin{align}
w &= \sum_{i=1}^{n}(\hat{\alpha_i}-\alpha_i)x_i \\
0  &= \sum_{i=1}^{n}(\hat{\alpha_i}-\alpha_i) \\
 C  &= \alpha_i + \mu_i \\
 C  &= \hat{\alpha_i} + \hat{\mu_i}
\end{align}
Thus, we can get Dual Problem for SVR.
\begin{equation} 
\begin{aligned}
\max_{\alpha,\hat{\alpha}} \, &  \sum_{i=1}^{n}\hat{y_i}(\hat{\alpha_i} - \alpha_i) - \epsilon(\hat{\alpha_i} + \alpha_i) \\
& -\frac{1}{2}\sum_{i=1}^{n}\sum_{j=1}^{n}(\hat{\alpha_i}-\alpha_i)(\hat{\alpha_j} - \alpha_j)(x_i)^Tx_j \\
s.t. & \, \sum_{i=1}^{n}(\hat{\alpha_i} - \alpha_i) = 0
\end{aligned}
\end{equation}
\end{itemize}
\begin{itemize}
\item Kernel Trick. Typically, the model is nonlinear in 1D space. Thus, we use $\phi(x)$ to represent the transformed x. Some common kernel functions are listed in table 3. \begin{align}
w &= \sum_{i=1}^{n}(\hat{\alpha_i} - \alpha_i)\phi(x)\\
f(x) &=  \sum_{i=1}^{n}(\hat{\alpha_i} - \alpha_i)k(x,x_i) + b
\end{align}

\begin{table}[!h]\centering \setlength{\belowcaptionskip}{0.1cm}
\caption{Common Kernel Functions}
\begin{tabular}{lll}
\toprule  
 $Name$ & $Expression$ & $Parameters$\\
\midrule  
Linear Kernel   & $\mathrm{k}\left(\mathrm{x}_{i}, x_{j}\right)=\left(x_{i}\right)^{T} x_{j}$ & d = 1\\
Polynomial Kernel & $\mathrm{k}\left(\mathrm{x}_{i}, x_{j}\right)=\left(\left(x_{i}\right)^{T} x_{j}\right)^{d}$& $\geq 1$\\
Gaussian Kernel &  $\mathrm{k}\left(\mathrm{x}_{i}, x_{j}\right)=\exp \left(-\frac{\left\|x_{i}-x_{j}\right\|^{2}}{2 \sigma^{2}}\right)$& $\sigma >0$ \\
Sigmoid Kernel & $\mathrm{k}\left(\mathrm{x}_{i}, x_{j}\right)=\tanh \left(\beta\left(x_{i}\right)^{T} x_{j}+\theta\right)$& $\beta > 0,\theta \textless 0$\\
\bottomrule 
 \end{tabular}
 \end{table} 
 \end{itemize}
 \subsection{Random Forest Regression}

\begin{itemize}
\item Tree Structure \cite{tree}. A walk-through of tree structure is going to be given in this section. In regression tree, there're root node,internal nodes and leaf node. For instance, if we're going to predict a man's height. And we get men's and women's heights. Thus, the root node is man or not. If a point which we need to predict is $(68,171)$. It means the height of a man weighing 68kg is 171cm. If we use tree regression, then our internal node is $>60?$ And the second internal node is $<70?$ and so on. And at last, the leaf node is the result we predict. Hence, the leaf node is the predicted height. 
\end{itemize} 
\begin{itemize}
    \item Classification And Regression Tree \cite{CART}. CART includes feature selection,generating trees and pruning. CART assumes that decision tree is a Binary tree. The decision tree is equivalent to dividing features into two groups recursively. It divides the input space into limited units and predict the distribution. Suppose our data $D = \lbrace(x_1,y_1),(x_2,y_2),\dots,(x_n,y_n)\rbrace$. Decision Tree splits input space into $M$ units $R_1,R_2,\dots,R_M$ and at the end of every unit is the output value $c_m$. And we can minimize squared errors of output values and true values. And it's obvious that the best output value in every unit should be the average value. But the question is that how to split the input space. And we choose $x_j$ randomly and its output value $s$. And we split the input space into two space by their output values. $R_1(j,s) = \lbrace x|x_j \leq s \rbrace$ and $R_2(j,s) = \lbrace x|x_j > s\rbrace$ $x_j$ is called splitting variable and $s$ is splitting point. And we find $c_1,c_2$ in $R_1,R_2$ via squared least errors. We also want $c_1 + c_2$ to be small enough. That's how we finally get j, s.

\begin{align}
    & \min \sum_{i=1} ^{n} (y_i - \hat{y_i})^2\\
    &\hat{c_m}  = average(y_i|x_i \in R_m)\\
    & \min_{j,s} (\min_{c1}\sum_{x_i \in R_1}(y_i - c_1)^2 + \min_{c_2}\sum_{x_i \in R_2}(y_i - c_2)^2)
\end{align}

\end{itemize}

\begin{itemize}
    \item Pruning \cite{CART}. Decision Tree implements recursive binary splitting to make more accurate predictions. And if the size of data and the input space is quite large. The structure of the tree is complicated. Sadly, a complex model easily gives birth to Overfitting. So it's necessary to make our model more simple. Therefore, from the bottom of the tree, we cut down some child trees. Then we can get a tree sequence $\lbrace T_0,T_1,\dots,T_n\rbrace$($T_0$ is the root node). Then we employ cross-validation to choose the best child tree from it. At the same time, we also expect our model to perform well. Hence, we apply a loss function to measure the differences of performances in the process of pruning. In equation(32), T is arbitrary child tree. C(T) means errors on the training data. $|T|$ is the number of leaf nodes in a child tree. $\alpha$ is a parameter, which decides the regularization term. If $\alpha$ is big, it means hard punishment and this results in a simple tree. If $\alpha$ is small, it means soft punishment and this leads to a more complicated tree respectively.

\begin{equation}
    C_\alpha(T) = C(T) + \alpha|T|
\end{equation}

\end{itemize}
\begin{itemize}
    \item Ensemble Learning \cite{random forest}. We use $1/3$ of our data to evaluate our model, which is called out of bag data. And $2/3$ of our data to be a new data set. Then we select subsets randomly from the new data set, which is known as Bagging. Every time we select one subset of the complete data set and then the subset is placed back. The number of points in different subsets is the same. We train different tree models for every different data. Finally we make an average of all trees' variables as our final model variable in order to cut down on the variance. Hence, it's an accurate model. And it is able to maintain accuracy although most of data is missing in that the model only randomly select a subset to train. However, it may overfit data when there're some outliers in data.
\end{itemize}

\subsection{Boosted Regression Tree}
\begin{itemize}
    \item Biased Feature Selection \cite{boosted}. It's almost the same as Random Forest. They both select the random subset and then a new tree is generated. In Random Forest, the probability of data is selected is the same, which is almost unbiased. But in Boosted Regression Tree, it is going to give weights to every data point. For instance, first time we select a subset and we build a tree model for it. Before this subset is placed back, prediction errors are calculated for every point. If the error is high, this point is likely to be given large weights, which indicates its probability of being selected is higher than others. To summarize, Boosted Regression Tree focuses on the errors and is going to mix it. Whereas, if there're many outliers in data, it just sucks. But it is robust to missing values just like Random Forest because they both select subsets to fit.
\end{itemize}
\subsection{Elastic Net Regression}

\begin{itemize}
\item Background: Multicollinearity. One of OLS's assumptions is that no multicollinearity. However, in multivariate regression, $X$ can be sometimes dependent. If this happens, OLS, Ridge and Lasso fail to play their part. 
\item Encourage Group Effect \cite{Elastic}. In multivariate regression, if some samples is correlated to each other, OLS is likely to take one sample, not caring which one is selected while strongly correlated samples are on the same boat in Elastic Net Regression. And it view them as a whole, which is called group effect. It does automatic variable selection and continuous shrinkage, and it select groups of correlated samples \cite{Elastic}, which is similar to clustering methods. However, this model doesn't reduce the variance and extra bias increases.

\item Generation Of Lasso And Ridge. Elastic Net is a middle ground between Ridge Regression and Lasso Regression. It mixes Lasso's loss function with Ridge's. It has the parameter r to control the mix ratio. If $r = 0$, it's Ridge. If $r = 1$, it's Lasso.
\item $N \gg P$. N is the number of samples and P is the number of features of every sample. When in multivariate regression, Elastic Net Regression can play a key role in $N \gg P$ cases.
\end{itemize}
\begin{align}
        J(\theta) &= \frac{1}{n}\sum \limits_{i} ^ {n} (y_i - \hat{y_i})^2 + r\lambda||\theta_i||+ \frac{1-r}{2}  \lambda||\theta _i||^2
\end{align}

\subsection{Least Angle Regression}
\begin{itemize}
    \item Joint Least Squares Direction \cite{LARS}. To begin with, the model selects the coefficient $\beta_j$ and calculate errors. When some other sample $x_k$ has more correlation with errors than $x_j$ has. Hence, the model increases $(\beta_j,\beta_k)$ in their joint least squares direction until $x_m$ has more correlation with errors. Thus, the model increases $(\beta_j,\beta_k,\beta_m)$ in their joint least squares direction. The model comes to its end until all samples in the model. Therefore, it can also settle samples' autocorrelation in high dimension. OLS is the special case of LARS. When LARS doesn't increase in joint least squares direction, the model becoms OLS. By the way, LARS is also powerful when $N \gg P$ just like Elastic Net Regression.
\end{itemize}

\begin{itemize}
    \item Sensitive To Outliers. LARS is a method which iteratively refit the errors. When there're outliers in data, then LARS doesn't make sense. 
\end{itemize}

\begin{itemize}
    \item Easily Modified. It's simple for LARS to join with other models such as Lasso. LARS-Lasso employs the Lasso's loss function and applies the LARS's method of coefficient selection.
\end{itemize}

 \subsection{RANSAC Regression}
 
 \begin{itemize}
     \item Background. Although Lasso is insensitive to a few outliers, Lasso doesn't work when data is filled with a large number of outliers,let alone OLS.
 \end{itemize}
 \begin{itemize}
     \item Random Sample Consensus Set \cite{RANSAC}. RANSAC is an iterative method. And it has a error threshold $\epsilon$. To begin with, it select a subset of the whole data and find a model to fit it. Then, use the model to test the rest of the data. If point' loss on the model is no more than $\epsilon$ , then add it to the consensus set, which is full of inliers. And this process is iterative. When iterative times is reached, the process comes to its end. RANSAC is going to make the number of points in consensus set as large as possible. 

\begin{algorithm}
\hspace*{5mm}

iteration times n\\
i = 0\\
 \While{i \textless n}{
  Randomly Select Inliers From Data\\
  Find A Model M To Fit\\
  Test Other data Via The Model\\
\If {points fit M} {
Inliers Set $\gets$ points  \\
}
\Else 
{Ouliers Set $\gets$ points\\
}
}
 \caption{Random Sample Consensus}
\end{algorithm}
 \end{itemize}
 \begin{itemize}
 \item  Voting Scheme. RANSAC is kind of like voting process. A subset of data claims its idea and then the rest of the data votes for the idea. In every independent process, there are two kinds of data. One agrees with the idea while the other is against it. And RANSAC is going to select a process where the number of supporters is max. Thus,RANSAC is biased. If the model is going to be robust to outliers, then there must be enough good features which vote for correct model and outliers can't vote consistently. 
 \end{itemize}
 \begin{itemize}
 \item  Disadvantages. When there're few outliers in the data, RANSAC can't make sense in that the difference between every process is little. Only when the data is heavily contaminated, RANSAC can play its part. Besides, the threshold must be set by hand, which requires users to decide specific threshold on different data. Last but not least, the cost of computation is high because it is an iterative method and the number of times required in the model is unknown.
 \end{itemize}
 
 \subsection{Theil-Sen Regression}
\begin{itemize}
\item Median Method \cite{Theil-Sen}. There're many data pairs used to calculate coefficient. $\theta = y_a - y_b / x_a - x_b$ And $\theta$ is the median of all $\theta s$. $b = y - \theta x$ b is also the median of $bs$. Hence, it's a nonparametric technique. However, complete computation leads to low speed. 
\end{itemize}
\begin{itemize}
    \item Breaking Point. In particular, Theil-Sen only can tolerate 29.3\% of data is outliers. And when the model is applied in high-dimensional regression, the rate is going to decrease.
\end{itemize}
\subsection{Huber Regression}

\begin{itemize}
    \item Huber Loss \cite{Huber}. It transforms its loss function when faced with different values. When values are large, which is of high possibility of being outliers, Huber turns their loss function into Linear Loss in order to minimize their influence on the model. $\delta$ serves as a threshold, deciding how large data is to need a linear loss. And it is fastest in three robust regression. 
    \begin{eqnarray}
\centering
     Huber Loss=
     \begin{cases}
 \frac{1}{2} a^2&,if |a|  \leq \delta\\
\delta (|a| - \frac{1}{2}\delta)&otherwise\\

\end{cases}
\end{eqnarray}
\end{itemize}

\subsection{Multivariate Adaptive Regression Splines}

\begin{itemize}
    \item Partitioning \cite{MARS}. MARS begins with partitioning data and then runs linear regression on each different partition. And it makes no assumptions about the relationship between the labels and samples. MARS originally has a large collection of basis functions. Each meeting point of two linear models is called a knot. And each knot has a pair of basis functions. And these functions are used to describe the relationship between $x$ and $y$. The first basis function is $max(0,x - y)$. The second is $max(0, y - x)$. 
\end{itemize}
\begin{itemize}
    \item Remove Basis Functions \cite{MARS}. After MARS partitions data and builds models, it applies least-squares model to fit data. And each knot has two basis functions. The results of them can be viewed as input variables. Least-Squares model estimate the loss of each basis function's output value. If a basis function has little influence on model fitting, then it is going to be removed. 
\end{itemize}

\begin{itemize}
    \item Advantages And Disadvantages. It can fit a large number of predictor variables. And it is an effective and fast algorithm. Also, it is robust to outliers. However, it begins with a large set of models and this easily leads to overfitting. And it is vulnerable to missing data problems. 
\end{itemize}
\subsection{Polynomial Regression}

\begin{itemize}
\item Background: When the relationship between $X$ and $y$ is nonlinear, OLS sucks.
\item Polynomial Transformation \cite{Polynomial}. Polynomial Regression replaces original $X$ with Polynomial in order to attain a more linear relationship than before or change features for some reason. Hence, it is not interpretable. Interestingly, it is somewhat like Talyor Extend. When your model breaks the assumption of linearity, then you can try all polynomial regression to find a best one, which is a good recipe.If feature's dimension is 2,
\end{itemize}
\begin{itemize}
    \item 
order = 1$\Rightarrow$ $[1,X_1,X_2]$
\end{itemize}
\begin{itemize}
    \item 
order = 2$\Rightarrow$ $[1,X_1,X_2,X_1^2,X_1X_2,X_2^2]$
\end{itemize}
\begin{itemize}
    \item 
 order = 3$\Rightarrow$ $[1,X_1,X_2,X_1 ^2,X_1X_2,X_2^2,X_1^3,X_1^2X_2,X_1X_2^2,X_2^3]$

\end{itemize}
\subsection{Weighted Least Squares}

\begin{itemize}
    \item Background. One of the OLS's assumptions is constant error variance. In section 3, I put forward log method. However, it is ineffective. 
\end{itemize}
\begin{itemize}
    \item Transformed Weights \cite{WLS}. When $w_i$ all equals 1, it should be the OLS. In OLS, the model gives every point the same attention. But it's under homoskedasticity while we come across more heteroskedastic scenarios. The idea is that we gives more attention to those points of which error is small. Thus, the model gives those points bigger weights.
    \end{itemize}
    \begin{itemize}
        \item Stable Intercept and Sensitive Coefficient \cite{WLS_2}. Note I choose different values for the first 20 weights and others are always 1. Different weights are equivalent for errors' abnormal distribution. As shown in Table 4, the intercept is right regardless. However, the coefficient changes sharply. Hence, we can't use it to draw inferences and test our hypotheses with regard to coefficient. 

\begin{equation}
    WLS = \sum\limits_{i=1}^ {n} w_i(y_i - \hat{y_i})^2
\end{equation}

 	\begin{figure}[htbp]
		\center
		\subfigure[w = 1]{\label{fig:1a}
		\includegraphics[width=0.3\linewidth]{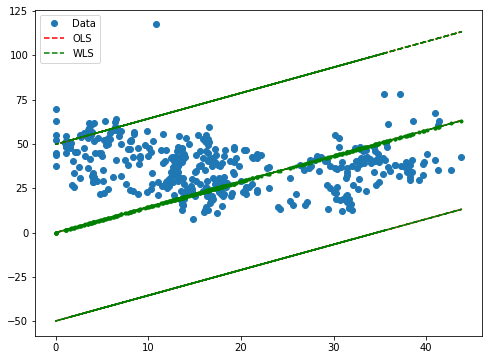}}
		\subfigure[w = 10]{\label{fig:1b}
		\includegraphics[width=0.3\linewidth]{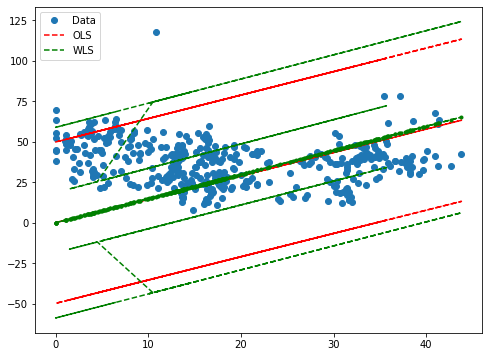}}
		\subfigure[w = 20]{\label{fig:1c}
		\includegraphics[width=0.3\linewidth]{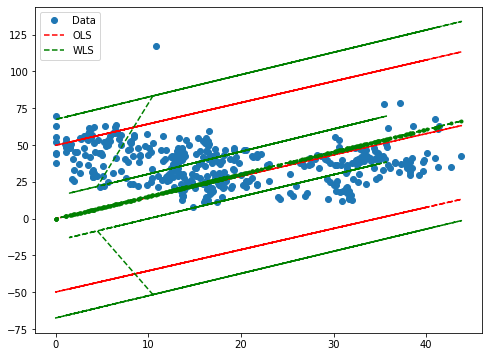}}
	\caption{The middle line shows OLS and MLS fits the data.And others show the range of predicted values of different algorithms}

	\end{figure}
	
\begin{table}[!h]
\centering
\setlength{\belowcaptionskip}{0.1cm}
\caption{Slopes And Intercepts}
\begin{tabular}{lll}
\toprule  
 $Weights$ & $Slopes$& $Intercepts$\\
\midrule  
1.0 & 1.444    & 0.059\\
10.0 &	1.4887 & 0.059	\\
20.0 & 1.5146	 & 0.058\\
40.0 & 1.5407    & 0.058 \\
80.0 & 1.5615  & 0.057\\
160.0&	1.5755  & 0.057\\
\bottomrule 
  
 \end{tabular}

 \end{table}
 As shown in Figure 9,Figure(a) shows that OLS is the special case of WLS when weights = 1.And Figure(b) and Figure(c) show when weights change,the intercept almost stays the same.
\end{itemize}

\subsection{Generalized Least Squares}
\begin{itemize}
    \item Background. OLS assumes that the error must be independent in the sense that one error can't be correlated to others. But when error autocorrelation happens, OLS has no way but to fail.
\end{itemize}
\begin{itemize}
    \item Better OLS \cite{GLS_1}. GLS is similar to OLS on a linearly transformed version of the data. And GLS is unbaised, consistent and effective. WLS is the special case of GLS, which means GLS can also solve heteroskedasticity \cite{GLS_2}.  
\end{itemize}

\subsection{Feasible Generalized Least Squares}
\begin{itemize}
    \item Implementable GLS. While GLS sounds powerful, but it can't be applied in specific regression tasks. FGLS is an implementable version of GLS. And FGLS needs some crucial assumptions to ensure a consistent estimator for errors covariance matrix.
\end{itemize}
\begin{itemize}
    \item Inefficiency On Little Data. Whereas GLS is more powerful than OLS under heteroscedasticity or autocorrelation, this is not the case for FGLS. When the size of data is quite small, FGLS is ineffctive than OLS. Thus, some people prefer FGLS over OLS under small data. But when the size of data becomes large, FGLS is a better choice.
\end{itemize}

\subsection{Bayesian Regression}

\begin{itemize}
    \item Background. As is known to all, OLS exactly makes an estimation of the mean of the values, which fails to provide a whole picture of the relationship between independent variables and dependent variables. And in some cases, we want to obtain a possible distribution of labels instead of a mean value.
\end{itemize}
\begin{itemize}
    \item Bayesian Theorem \cite{bay}. For instance, we're going to employ a model to distinguish whether a email is normal or spam. So what our model faces is that it has to make predictions about the unknown email. Our data has 100 emails and 10\% of them is spam. Hence, the percentage of spam is 10\%. But that's absolutely not the whole story. In Bayesian, it's called Prior Probability, which means the Basic Assumption of the distribution and that's where Bayesian Begins. At the beginning of the algorithm, Bayesian is biased in return the model is easily affected by the original distribution. For example, if we only have all 10 normal emails, it's impossible that we wouldn't get any spam emails in future. In other words, if the size of our data is quite small, it's not incentive for us to implement Bayesian. However, when training times keep increasing, we should get ideal results ultimately. In the equation below, P(B) is a Normalization term and P(A) is Prior Probability. $P(A|B)$ is called  Posterior Probability(Conditional Probability). To conclude, when we have much data, Bayesian may be a good choice to try while it exactly performs like other algorithms.
\end{itemize}

\begin{align}
    P(A|B) &= \frac{P(B|A) P(A)}{P(B)}
    \end{align}

\begin{itemize}
    \item Maximum Likelihood Estimation \cite{bay_2}.
Generally speaking, our goal is to figure out the real data distribution, which is almost impossible. Therefore, we want a data distribution which is close to our data distribution from a problem domain. MLE(Maximum Likelihood Estimation) indicates that we want to maximize the probability that real data is sampled from the Hypothesis Distribution.
\end{itemize}
\begin{equation}
    \beta^* = \arg \max\limits_{\beta} P_\beta(D)
\end{equation}

\begin{itemize}
    \item Maximum Posterior Estimation \cite{bay_3}.
Typically, we can use MAP(Maximum A Posterior Estimation) to replace MLE. It's based on Bayesian Theorem. And MAP is fundamental to Bayesian Regression(equation 37). Rather than other standard algorithms, Bayesian Regression doesn't produce a single value but a range of possible distribution. And in most cases, MLE and MAP are likely to get the same results. However, when the hypothesis of MAP is different from MLE, they fail to reach the same destination. When Prior Probability is uniformly distributed, they can make it. From another point of view, if we have some precise understanding of data, Bayesian Regression is a excellent choice in that it serves as Prior Probablity or we can weigh every different choice just like Weighted Least Errors. Interestingly, prior can be kind of Regularization or bias of the model, as such prior can be interpreted as L2 norm, which is also called Bayesian Ridge Regression. Equation(38) means given a model $m$, the probability of output y. And $\beta$(Coefficients) and $\sigma$(Standard Deviation) are arbitrary values.

\end{itemize}
\begin{align}
    P(\beta|D) &= \frac{P(D|\beta)P(\beta)}{P(D)}\\
    P(y|m) &= \frac{P(\beta,\sigma|m)P(y|X,\beta,\sigma,m)}{P(\beta,\sigma|y,X,m)}\\
    y &\sim (\beta^TX,\sigma^2)
\end{align}

\subsection{Quantile Regression}

\begin{itemize}
    \item Transformed Loss Function \cite{QR}. QR has a parameter q which decides the proportion to split the data. One is q \% of the data and the other is (1-q)\% of the data. For instance, if q = 0.5, then data is split in two. We minimize the squared loss in OLS while we now minimize the absolute loss in QR.
\end{itemize}
\begin{equation}
    J(\theta) =  \sum_{i=1}^{n}q|y_i - \hat{y_i}| + \sum_{i=1}^{n}(1-q)|y_i - \hat{y_i}|
\end{equation}

\begin{itemize}
    \item Advantages. It goes without saying that QR can provide a more complete view of the relationship than OLS. What's more, it is also robust to outliers and situations where the variance of errors is not a constant.
\end{itemize}

\subsection{Ordinal Regression}

\begin{itemize}
    \item Background. In some cases, the values for labels are ranking numbers. For instance, 0-5 can represent his ability of communicating with others in social science. And OLS can't make accurate prediction about them.
\end{itemize}

\begin{itemize}
    \item Ranking Learning \cite{OR}. In OR, the model has a set of thresholds $\theta_1,\theta_2,\dots,\theta_n$, which is used to split predictions into independent intervals and every interval corresponds to a $y$. The model can be represented by sigmoid function of which output values stand for possibility. 
\end{itemize}
\begin{equation}
    P(y\leq i|x) = \sigma(\theta_i - \hat{y_i})
\end{equation}
\section{Extra Models}

\subsection{Generalized Linear Models}

\begin{itemize}
    \item Generalized Functions \cite{GLM}. Just as the name implies, it is generalization of different functions. And it consists of two significant parts. The first part is the probability distribution of $y$ such as normal distribution(OLS). The second is linear predictor, which decides how the coefficients combine with independent variables. And GLM includes several regression models such as Binomial Regression, Bernoulli Regression, Poisson Regression and so on. Their application is not so wide and they just swift the two parts compared to OLS. Hence, it's left out in this paper.
\end{itemize}

\begin{itemize}
    \item Advantages And Disadvantages. It can absolutely deal with situations where $y$ doesn't follow normal distribution. However, it needs large data sets and it is sensitive to outliers.
\end{itemize}
\subsection{Step-Wise Regression}

\begin{itemize}
    \item Forward Selection \cite{Step_1}. This method begins with no variables. And it involves testing the addition of the variable in an iterative method which is of great use to the improvement of the accuracy. The model repeats until no improvement.  
\end{itemize}

\begin{itemize}
    \item Backward Elimination. This method starts with many candidate variables. It involves testing the loss of the model with the deletion of variables. If the loss is small, then the variable is going to be deleted. The model repeats until no variable can be deleted.
\end{itemize}
\begin{itemize}
    \item Bidirectional Elimination. This is an combination of the above two methods. Whether adding or deleting a variable is decided on every step. To summarize, Step-Wise Regression contains a big space of possible models, which can lead to overfitting.
\end{itemize}

\begin{itemize}
    \item Reasons For Stopping \cite{Step_2}. First, the tests such as F-tests and t-tests are biased, thus it may not be accurate. Second, widespread incorrect usage and availability of alternative models such as ensemble learning have led to calls to stop the use of this algorithm.
\end{itemize}

\section{Relationship With Deep Learning}

\subsection{General Regression Neural Network}
\begin{figure}[htbp]
\includegraphics[height=2in]{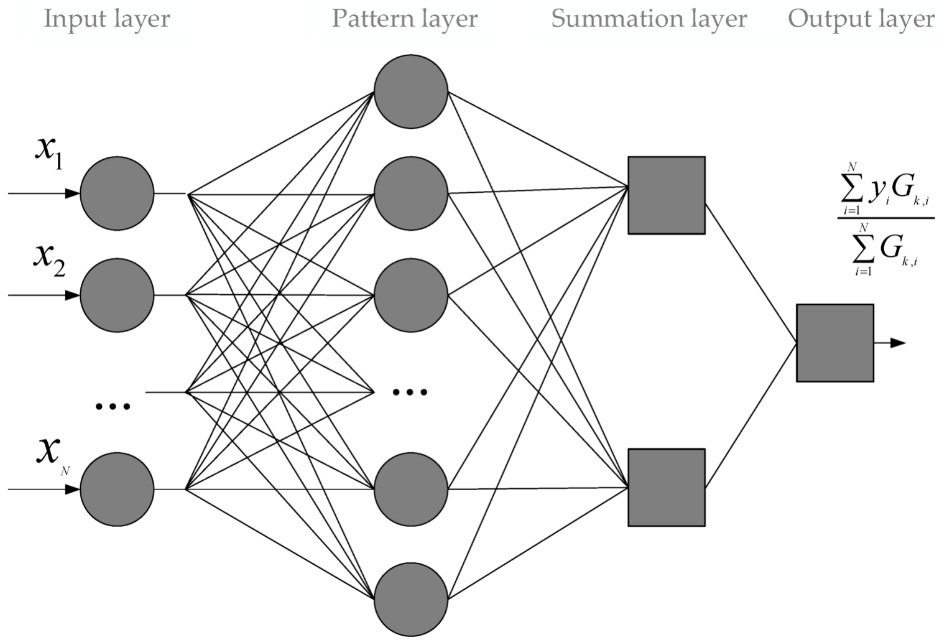}
\caption{GRNN Structure. Source: https://www.mdpi.com/1424-8220/20/9/2625.}
\end{figure}

\begin{itemize}
    \item Network Structure \cite{GRNN}. GRNN includes input, pattern, summation and output layers. The input and output layers are independent vector and dependent vector. The pattern layer can be seen as a vector full of coefficients. For instance, if we want to apply $y = \theta x$. Then one pattern neuron stands for $\theta_i x_i$. And output layer can be formulated as the equation below. And this model can maintain its accuracy with small data and it's robust to outliers. However, the structure of the network is complicated so that it is computationally expensive.
\end{itemize}

\begin{align}
    &Y(x) = \frac{\sum_{i=1}^{n}y_iK(x,x_i)}{\sum_{i=1}^{n}K(x,x_i)}\\
    &K(x,x_i) = e^{-d_i/2\sigma^2} \\
    &d_i = (x - x_i)^T(x - x_i)
\end{align}

\begin{itemize}
    \item Widespread Application \cite{GRNN_2}. Many regression models like Poisson Regression and Ordinal Regression have succeeded in using GRNN. And we can draw a safe conclusion that a complicated network structure can represent any kind of regression. But only few of them are proved successful. Neural Network is powerful and classic regression algorithms are well-structured. Maybe regression can be applied in neural network without missing its original function. Humans have made fundamental progress in Regression. If we can combine Regression with neural network perfectly, then it's another picture.
\end{itemize}

\section{Conclusions}

In this paper, I set out necessary assumptions with OLS and little tricks to fix the problems when assumptions are violated. Amazingly, it seems that OLS is the beginning of almost evey regression model. And a large number of Regression models are designed to be a better OLS. They can play their part in situations where OLS fails to work. I hold the belief that not evey algorithm needs to be introduced in details. Hence, the widespread algorithms are given enough attention and others are quickly illustrated. Finally, I give a quick overview of GRNN. From this paper, I can draw three conclusions.

\begin{itemize}
    \item Know Your Model. Note that regression algorithms aren't plug-and-play. You must know evey model's range of application and are able to deal with situations where the model's assumptions are unsatisfied.
\end{itemize}

\begin{itemize}
    \item Regression In the Future. Regression is older compared to Deep Learning and great ideas behind every classic algorithm is never out of date. And people always want to predict unknown values and regression task is really fascinating. Deep Learning is quite powerful. If regression can learn from Deep Learning and keeps its excellent part, I do believe regression can be more powerful in the near future.
\end{itemize}
\begin{itemize}
    \item It seems that regression algorithms are out of date. However, as far as I am concerned, beautiful ideas behind every algorithm are shared. In other words, dipping into these old algorithms can enable us to gain insight and intuition about algorithms and put forward exciting algorithms which share the same ideas with regression algorithms and are just different implementations of awesome ideas to handle new problems.
\end{itemize}


\begin{thebibliography}{9}
\bibitem{Legendre}
A.M. Legendre. Nouvelles méthodes pour la détermination des orbites des comètes, Firmin Didot, Paris, 1805. “Sur la Méthode des moindres quarrés” appears as an appendix.

\bibitem{Gauss}
C.F. Gauss. Theoria combinationis observationum erroribus minimis obnoxiae

\bibitem{Ridge_1}
Arthur E. Hoerl \& Robert W. Kennard (1970) Ridge Regression: Biased Estimation for Nonorthogonal Problems, Technometrics, 12:1, 55-67, DOI: 10.1080/00401706.1970.10488634

\bibitem{Ridge_2}
Arthur E. Hoerl \& Robert W. Kennard (1970) Ridge Regression: Applications to Nonorthogonal Problems, Technometrics, 12:1, 69-82, DOI: 10.1080/00401706.1970.10488635

\bibitem{Lasso}
Tibshirani, R. (1996), Regression Shrinkage and Selection Via the Lasso. Journal of the Royal Statistical Society: Series B (Methodological), 58: 267-288. https://doi.org/10.1111/j.2517-6161.1996.tb02080.x


\bibitem{Elastic}
Zou, H. and Hastie, T. (2005), Regularization and variable selection via the elastic net. Journal of the Royal Statistical Society: Series B (Statistical Methodology), 67: 301-320. https://doi.org/10.1111/j.1467-9868.2005.00503.x

\bibitem{Hetero}
Hayashi, Fumio (2000). Econometics. Princeton University Press. p. 15.

\bibitem{autocorrelation}
 Gubner, John A. (2006). Probability and Random Processes for Electrical and Computer Engineers. Cambridge University Press. ISBN 978-0-521-86470-1.
 
\bibitem{svr_1}
 V. Vapnik, “The support vector method of function estimation, ” in J.A.K. Suykens and J. Vandewalle (Eds) Nonlinear Modeling: Advanced Black-Box Techniques, Kluwer Academic Publishers, Boston, pp. 55–85, 1998.
 
\bibitem{svr_2}
 Drucker, Harris; Burges, Christ. C.; Kaufman, Linda; Smola, Alexander J.; and Vapnik, Vladimir N. (1997); "Support Vector Regression Machines", in Advances in Neural Information Processing Systems 9, NIPS 1996, 155–161, MIT Press.
 
\bibitem{svr_3}
Smola, A.J., Schölkopf, B. A tutorial on support vector regression. Statistics and Computing 14, 199–222 (2004). https://doi.org/10.1023/B:STCO.0000035301.49549.88

\bibitem{tree}
S. R. Safavian and D. Landgrebe, "A survey of decision tree classifier methodology," in IEEE Transactions on Systems, Man, and Cybernetics, vol. 21, no. 3, pp. 660-674, May-June 1991, doi: 10.1109/21.97458.

\bibitem{CART}
Loh, W.‐Y. (2011), Classification and regression trees. WIREs Data Mining Knowl Discov, 1: 14-23. https://doi.org/10.1002/widm.8

\bibitem{random forest}
Liaw A, Wiener M. Classification and regression by randomForest[J]. R news, 2002, 2(3): 18-22.

\bibitem{boosted}
Elith J, Leathwick J R, Hastie T. A working guide to boosted regression trees[J]. Journal of Animal Ecology, 2008, 77(4): 802-813.

\bibitem{LARS}
 Efron, Bradley; Hastie, Trevor; Johnstone, Iain; Tibshirani, Robert (2004). "Least Angle Regression" (PDF). Annals of Statistics. 32 (2): pp. 407–499. arXiv:math/0406456. doi:10.1214/009053604000000067. MR 2060166.
 
\bibitem{RANSAC}
Fischler, M. \& Bolles, R. ( 1981). Random Sample Consensus: A Paradigm for Model Fitting with Applications to Image Analysis and Automated Cartography. Communications of the ACM, 24, 381-395.

\bibitem{Theil-Sen}
Theil, H. (1950), "A rank-invariant method of linear and polynomial regression analysis. I, II, III", Nederl. Akad. Wetensch., Proc., 53: 386–392, 521–525, 1397–1412, MR 0036489

\bibitem{Huber}
Huber, Peter J. (1964). "Robust Estimation of a Location Parameter". Annals of Statistics. 53 (1): 73–101. doi:10.1214/aoms/1177703732. JSTOR 2238020.

\bibitem{MARS}
Friedman J H. Multivariate adaptive regression splines[J]. The annals of statistics, 1991: 1-67.

\bibitem{Polynomial}
Stigler, Stephen M. (November 1974). "Gergonne's 1815 paper on the design and analysis of polynomial regression experiments". Historia Mathematica. 1 (4): 431–439. doi:10.1016/0315-0860(74)90033-0.

\bibitem{WLS}
Ruppert D, Wand M P. Multivariate locally weighted least squares regression[J]. The annals of statistics, 1994: 1346-1370.

\bibitem{WLS_2}
Suykens J A K, De Brabanter J, Lukas L, et al. Weighted least squares support vector machines: robustness and sparse approximation[J]. Neurocomputing, 2002, 48(1-4): 85-105.

\bibitem{GLS_1}
Amemiya, Takeshi (1985). "Generalized Least Squares Theory". Advanced Econometrics. Harvard University Press. ISBN 0-674-00560-0.

\bibitem{GLS_2}
Kmenta, Jan (1986). "Generalized Linear Regression Model and Its Applications". Elements of Econometrics (Second ed.). New York: Macmillan. pp. 607–650. ISBN 0-472-10886-7.

\bibitem{GLS_3}
Kariya T, Kurata H. Generalized least squares[M]. John Wiley \& Sons, 2004.

\bibitem{bay}
Bernardo J M, Smith A F M. Bayesian theory[M]. John Wiley \& Sons, 2009.

\bibitem{bay_2}
Myung I J. Tutorial on maximum likelihood estimation[J]. Journal of mathematical Psychology, 2003, 47(1): 90-100.

\bibitem{bay_3}
Gauvain J L, Lee C H. Maximum a posteriori estimation for multivariate Gaussian mixture observations of Markov chains[J]. IEEE transactions on speech and audio processing, 1994, 2(2): 291-298.

\bibitem{QR}
Koenker R, Hallock K F. Quantile regression[J]. Journal of economic perspectives, 2001, 15(4): 143-156.


\bibitem{OR}
Harrell Jr F E. Regression modeling strategies: with applications to linear models, logistic and ordinal regression, and survival analysis[M]. Springer, 2015.

\bibitem{GLM}
Faraway J J. Extending the linear model with R: generalized linear, mixed effects and nonparametric regression models[M]. CRC press, 2016.

\bibitem{Step_1}
Efroymson, MA (1960) "Multiple regression analysis." In Ralston, A. and Wilf, HS, editors, Mathematical Methods for Digital Computers. Wiley.

\bibitem{Step_2}
Flom, P. L. and Cassell, D. L. (2007) "Stopping stepwise: Why stepwise and similar selection methods are bad, and what you should use," NESUG 2007.

\bibitem{GRNN}
Specht, D. F. (2002-08-06). "A general regression neural network". IEEE Transactions on Neural Networks. 2 (6): 568–576. doi:10.1109/72.97934. PMID 18282872.

\bibitem{GRNN_2}
Dreiseitl S, Ohno-Machado L. Logistic regression and artificial neural network classification models: a methodology review[J]. Journal of biomedical informatics, 2002, 35(5-6): 352-359.

\bibitem{l}
Hutcheson G D. Ordinary least-squares regression[J]. L. Moutinho and GD Hutcheson, The SAGE dictionary of quantitative management research, 2011: 224-228.

\bibitem{l}
Dismuke C, Lindrooth R. Ordinary least squares[J]. Methods and Designs for Outcomes Research, 2006, 93: 93-104.

\bibitem{l}
Kiers H A L. Weighted least squares fitting using ordinary least squares algorithms[J]. Psychometrika, 1997, 62(2): 251-266.

\bibitem{l}
Marquardt D W, Snee R D. Ridge regression in practice[J]. The American Statistician, 1975, 29(1): 3-20.
\bibitem{l}
Hoerl A E, Kannard R W, Baldwin K F. Ridge regression: some simulations[J]. Communications in Statistics-Theory and Methods, 1975, 4(2): 105-123.
\bibitem{l}
Le Cessie S, Van Houwelingen J C. Ridge estimators in logistic regression[J]. Journal of the Royal Statistical Society: Series C (Applied Statistics), 1992, 41(1): 191-201.
\bibitem{l}
Osborne M R, Presnell B, Turlach B A. On the lasso and its dual[J]. Journal of Computational and Graphical statistics, 2000, 9(2): 319-337.
\bibitem{l}
Zou H. The adaptive lasso and its oracle properties[J]. Journal of the American statistical association, 2006, 101(476): 1418-1429.
\bibitem{l}
Park T, Casella G. The bayesian lasso[J]. Journal of the American Statistical Association, 2008, 103(482): 681-686.
\bibitem{l}
Zhao P, Yu B. On model selection consistency of Lasso[J]. The Journal of Machine Learning Research, 2006, 7: 2541-2563.
\bibitem{l}
Meinshausen N. Relaxed lasso[J]. Computational Statistics \& Data Analysis, 2007, 52(1): 374-393.
\bibitem{l}
Zou H, Hastie T. Regression shrinkage and selection via the elastic net, with applications to microarrays[J]. JR Stat Soc Ser B, 2003, 67: 301-20.
\bibitem{l}
Ogutu J O, Schulz-Streeck T, Piepho H P. Genomic selection using regularized linear regression models: ridge regression, lasso, elastic net and their extensions[C]//BMC proceedings. BioMed Central, 2012, 6(2): 1-6.

\bibitem{l}
Ceperic E, Ceperic V, Baric A. A strategy for short-term load forecasting by support vector regression machines[J]. IEEE Transactions on Power Systems, 2013, 28(4): 4356-4364.
\bibitem{l}
Angiulli G, Cacciola M, Versaci M. Microwave devices and antennas modelling by support vector regression machines[J]. IEEE Transactions on Magnetics, 2007, 43(4): 1589-1592.
\bibitem{l}
Xu S, An X, Qiao X, et al. Multi-output least-squares support vector regression machines[J]. Pattern Recognition Letters, 2013, 34(9): 1078-1084.
\bibitem{l}
Segal M R. Machine learning benchmarks and random forest regression[J]. 2004.
\bibitem{l}
Svetnik V, Liaw A, Tong C, et al. Random forest: a classification and regression tool for compound classification and QSAR modeling[J]. Journal of chemical information and computer sciences, 2003, 43(6): 1947-1958.
\bibitem{l}
Cootes T F, Ionita M C, Lindner C, et al. Robust and accurate shape model fitting using random forest regression voting[C]//European Conference on Computer Vision. Springer, Berlin, Heidelberg, 2012: 278-291.
\bibitem{l}
Jöreskog K G, Goldberger A S. Factor analysis by generalized least squares[J]. Psychometrika, 1972, 37(3): 243-260.
\bibitem{l}
Orsini N, Bellocco R, Greenland S. Generalized least squares for trend estimation of summarized dose–response data[J]. The stata journal, 2006, 6(1): 40-57.
\bibitem{l}
Browne M W. Generalized least squares estimators in the analysis of covariance structures[J]. South African statistical journal, 1974, 8(1): 1-24.
\bibitem{l}
Hao L, Naiman D Q, Naiman D Q. Quantile regression[M]. Sage, 2007.
\bibitem{l}
Yu K, Lu Z, Stander J. Quantile regression: applications and current research areas[J]. Journal of the Royal Statistical Society: Series D (The Statistician), 2003, 52(3): 331-350.
\bibitem{l}
Meinshausen N, Ridgeway G. Quantile regression forests[J]. Journal of Machine Learning Research, 2006, 7(6).
\bibitem{l}
Bishop C M, Tipping M E. Bayesian regression and classification[J]. Nato Science Series sub Series III Computer And Systems Sciences, 2003, 190: 267-288.
\bibitem{l}
Gelman A, Goodrich B, Gabry J, et al. R-squared for Bayesian regression models[J]. The American Statistician, 2019.
\bibitem{l}
Koop G M. Bayesian econometrics[M]. John Wiley \& Sons Inc., 2003.
\bibitem{l}
Yu K, Moyeed R A. Bayesian quantile regression[J]. Statistics \& Probability Letters, 2001, 54(4): 437-447.
\bibitem{l}
Willett J B, Singer J D. Another cautionary note about R 2: Its use in weighted least-squares regression analysis[J]. The American Statistician, 1988, 42(3): 236-238.
\bibitem{l}
Chang P T, Lee E S. A generalized fuzzy weighted least-squares regression[J]. Fuzzy Sets and Systems, 1996, 82(3): 289-298.
\bibitem{l}
Blatman G, Sudret B. Adaptive sparse polynomial chaos expansion based on least angle regression[J]. Journal of computational Physics, 2011, 230(6): 2345-2367.
\bibitem{l}
Khan J A, Van Aelst S, Zamar R H. Robust linear model selection based on least angle regression[J]. Journal of the American Statistical Association, 2007, 102(480): 1289-1299.
\bibitem{l}
Hesterberg T, Choi N H, Meier L, et al. Least angle and l1 penalized regression: A review[J]. Statistics Surveys, 2008, 2: 61-93.
\bibitem{l}
Christensen R H B. ordinal—regression models for ordinal data[J]. R package version, 2015, 28: 2015.
\bibitem{l}
Elith J, Leathwick J. Boosted Regression Trees for ecological modeling[J]. R Documentation. Available online: https://cran. r-project. org/web/packages/dismo/vignettes/brt. pdf (accessed on 12 June 2011), 2017.
\bibitem{l}
Tyree S, Weinberger K Q, Agrawal K, et al. Parallel boosted regression trees for web search ranking[C]//Proceedings of the 20th international conference on World wide web. 2011: 387-396.
\bibitem{l}
Choi S, Kim T, Yu W. Performance evaluation of RANSAC family[J]. Journal of Computer Vision, 1997, 24(3): 271-300.
\bibitem{l}
Derpanis K G. Overview of the RANSAC Algorithm[J]. Image Rochester NY, 2010, 4(1): 2-3.
\bibitem{l}
Wilcox R. A note on the Theil‐Sen regression estimator when the regressor is random and the error term is heteroscedastic[J]. Biometrical Journal: Journal of Mathematical Methods in Biosciences, 1998, 40(3): 261-268.
\bibitem{l}
Fernandes R, Leblanc S G. Parametric (modified least squares) and non-parametric (Theil–Sen) linear regressions for predicting biophysical parameters in the presence of measurement errors[J]. Remote Sensing of Environment, 2005, 95(3): 303-316.
\bibitem{l}
Sun Q, Zhou W X, Fan J. Adaptive huber regression[J]. Journal of the American Statistical Association, 2020, 115(529): 254-265.
\bibitem{l}
Fox J, Weisberg S. Robust regression[J]. An R and S-Plus companion to applied regression, 2002, 91.
\bibitem{l}
Ostertagová E. Modelling using polynomial regression[J]. Procedia Engineering, 2012, 48: 500-506.
\bibitem{l}
Theil H. A rank-invariant method of linear and polynomial regression analysis[M]//Henri Theil’s contributions to economics and econometrics. Springer, Dordrecht, 1992: 345-381.
\bibitem{l}
Bendel R B, Afifi A A. Comparison of stopping rules in forward “stepwise” regression[J]. Journal of the American Statistical association, 1977, 72(357): 46-53.
\bibitem{l}
Zheng B, Agresti A. Summarizing the predictive power of a generalized linear model[J]. Statistics in medicine, 2000, 19(13): 1771-1781.
\bibitem{l}
Graybill F A. Theory and application of the linear model[M]. North Scituate, MA: Duxbury press, 1976.

\end{thebibliography}
\end{document}